\documentclass[letterpaper]{article} 
\usepackage{aaai2026}  
\usepackage{times}  
\usepackage{helvet}  
\usepackage{courier}  
\usepackage[hyphens]{url}  
\usepackage{graphicx} 
\usepackage{natbib}  
\usepackage{caption} 
\usepackage{algorithm}
\usepackage{algorithmic}
\usepackage{booktabs}
\usepackage{multirow} 
\usepackage{amssymb}
\usepackage{amsmath} 
\usepackage{newfloat}
\usepackage{listings}
\DeclareCaptionStyle{ruled}{labelfont=normalfont,labelsep=colon,strut=off} 
\floatstyle{ruled}
\newfloat{listing}{tb}{lst}{}
\floatname{listing}{Listing}
\title{Exploring Category-level Articulated Object Pose Tracking on SE(3) Manifolds}

\author {
    Xianhui Meng\textsuperscript{\rm 1}\thanks{Xianhui Meng and Yukang Huo contributed equally. \\
    Corresponding authors$^\dagger$: Li Zhang and Jun Liu.},
    Yukang Huo\textsuperscript{\rm 2 $*$}, 
    Li Zhang\textsuperscript{\rm 1, 5 $\dagger$}, 
    Liu Liu\textsuperscript{\rm 3},
    Haonan Jiang\textsuperscript{\rm 7}, 
    Yan Zhong\textsuperscript{\rm 4}, \\
    Pingrui Zhang\textsuperscript{\rm 6, 8}, 
    Cewu Lu\textsuperscript{\rm 9 },
    Jun Liu\textsuperscript{\rm 1 $\dagger$}
}
\affiliations {
    \textsuperscript{\rm 1}Department of Electronic Engineering and Information Science, University of Science and Technology of China. China\\
    \textsuperscript{\rm 2}China Agricultural University. China. \textsuperscript{\rm 3}Hefei University of Technology, Hefei, China. \textsuperscript{\rm 4}Peking University. China.\\
    \textsuperscript{\rm 5}Hefei Institute of Physical Science, Chinese Academy of Sciences, China. \textsuperscript{\rm 6}Fudan university. China.\\
    \textsuperscript{\rm 7}Zhejiang University of Technology, Zhejiang, China.    
    \textsuperscript{\rm 8}Shanghai AI Lab. China.
    \textsuperscript{\rm 9}Shanghai Jiao Tong University. China\\
    \texttt{\{mengxh, zanly20\}@mail.ustc.edu.cn, yukanghuo.ai@gmail.com, junliu@ustc.edu.cn.}
}

\usepackage{bibentry}

\begin{document}

\maketitle

\begin{abstract}
Articulated objects are prevalent in daily life and robotic manipulation tasks. However, compared to rigid objects, pose tracking for articulated objects remains an underexplored problem due to their inherent kinematic constraints. To address these challenges, this work proposes a novel point-pair-based pose tracking framework, termed \textbf{PPF-Tracker}. The proposed framework first performs quasi-canonicalization of point clouds in the SE(3) Lie group space, and then models articulated objects using Point Pair Features (PPF) to predict pose voting parameters by leveraging the invariance properties of SE(3). Finally, semantic information of joint axes is incorporated to impose unified kinematic constraints across all parts of the articulated object. PPF-Tracker is systematically evaluated on both synthetic datasets and real-world scenarios, demonstrating strong generalization across diverse and challenging environments. Experimental results highlight the effectiveness and robustness of PPF-Tracker in multi-frame pose tracking of articulated objects. We believe this work can foster advances in robotics, embodied intelligence, and augmented reality. Codes are available at \url{https://github.com/mengxh20/PPFTracker}.
\end{abstract}


\section{Introduction}

Domestic service robots must perceive and understand diverse 3D objects in complex human-centric environments~\cite{mo2021where2act}. Articulated objects (e.g., cabinets with doors and drawers) pose particular challenges due to structural complexity and semantic richness. Accurate six-degree-of-freedom (6-DoF) pose estimation~\cite{wen2023bundlesdf} is fundamental to downstream applications in embodied intelligence, VR/AR~\cite{davis2024analyzing,zhao2024wavelet}, human–computer interaction~\cite{clark2022articulate,Yang_2022_CVPR}, and robotic manipulation~\cite{jain2021screwnet,zhao2025ultrahr}. Reliable 6-DoF pose enables robust interpretation and interaction, supporting sophisticated tasks in dynamic, unstructured environments.

\begin{figure}[t!]
    \centering
    \includegraphics[width=\linewidth]{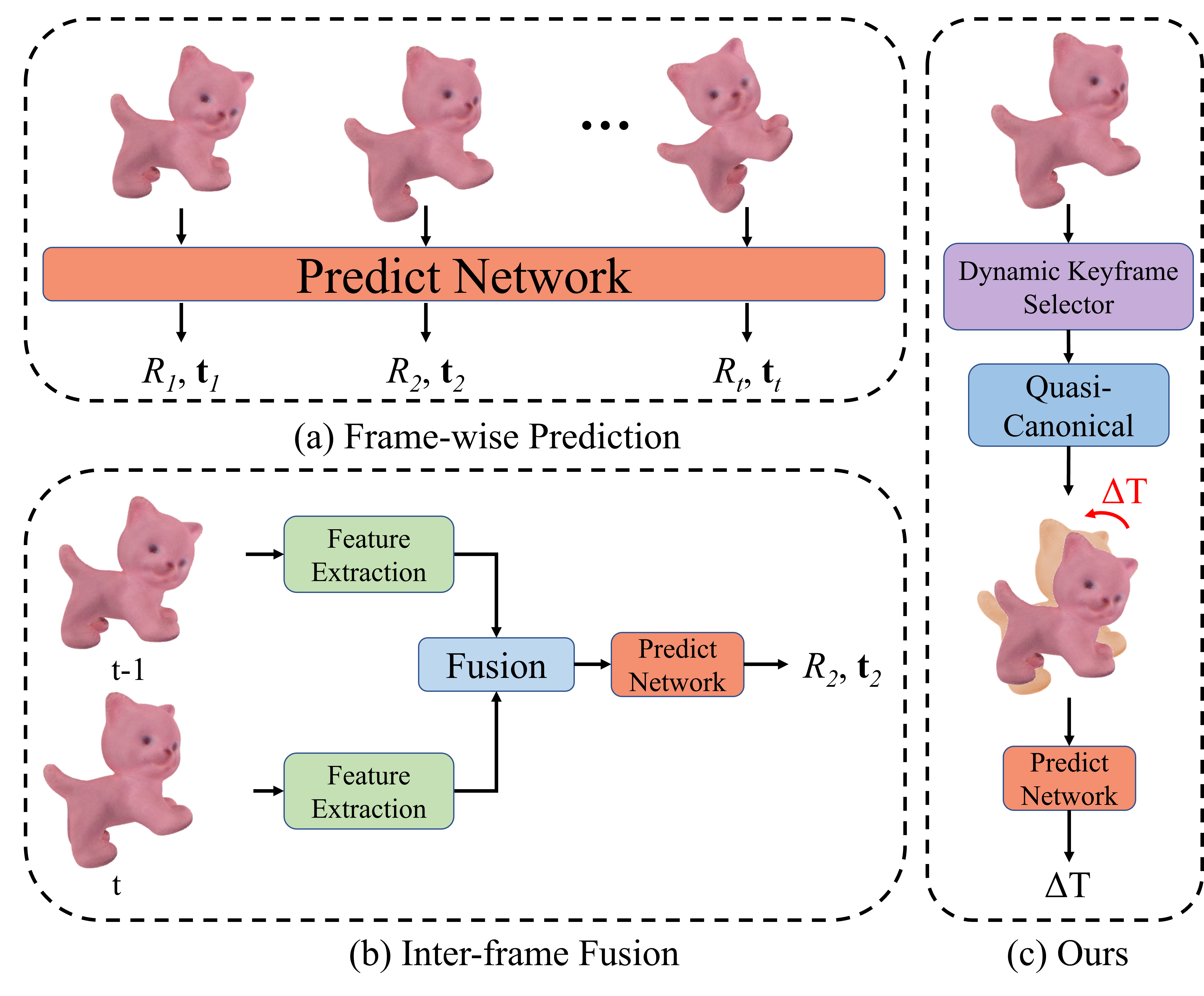}
    \caption{The Categorization of Tracking Methods.} 
    \label{fig:teaser}
\end{figure} 

Compared to instance-level tasks~\cite{mao2021fcpose,wen2021bundletrack,wang2018sgpn}, category-level 6D object pose estimation presents significantly greater challenges~\cite{heppert2022category,fu2022category,zhao2025zero,zhao2024cycle}. In category-level scenarios, models must estimate the 3D rotation and translation of previously unseen objects based on partial observations, without access to specific CAD models or instance-level priors. While recent methods~\cite{zhang2024generative,zou2024learning} have made encouraging progress in this domain, several critical challenges remain:

(1) \textbf{Pose Invalidity and Singularity}. Many existing methods optimize SE(3)-related parameters (e.g., Euler angles or translations) in Euclidean space, which can lead to invalid rotation matrices that violate orthogonality constraints. Moreover, Euler angles suffer from gimbal lock, while quaternions have sign ambiguity, both of which can cause singularities and unstable pose predictions.

(2) \textbf{Tracking Methods}. Traditional approaches often depend on dense predictions or frame-by-frame pose updates (Fig.~\ref{fig:teaser} (a,b))~\cite{you2020multi,wang20123}, which are computationally costly and unsuitable for real-time point cloud streams. These limitations hinder robust and continuous tracking in dynamic environments.

To address the first challenge, we represent object poses using the Lie group \( \mathrm{SE}(3) \) and perform optimization in its tangent space \( \mathfrak{se}(3) \), ensuring geometric consistency. Based on this formulation, we design a framework that leverages SE(3)-invariance for robust pose tracking. We use \textbf{Point Pair Features (PPF)} to encode relative pose information, and introduce a \textbf{kinematics-aware optimization module} to enforce structural constraints among parts, improving overall accuracy and consistency.

To address the second challenge, we incorporate temporal priors from adjacent frames to guide pose prediction (Fig. ~\ref{fig:teaser}(c)). At each time step, features are extracted and object states updated using information from previous frames. This inter-frame fusion improves tracking stability and reduces computational overhead, enabling efficient real-time performance on continuous point cloud streams.

We conduct a comprehensive evaluation of the proposed \textbf{PPF-Tracker} across multiple datasets. The results demonstrate that PPF-Tracker achieves strong performance on both synthetic dataset (PM-Videos) and semi-synthetic dataset (ReArt-Videos). Additionally, its consistent success in real-world scenarios (RobotArm-Videos) further highlights the method’s robustness and high generalization capability across diverse environments, validating its effectiveness in practical applications. The key contributions of this work are summarized as follows:

\begin{itemize}
    \item We propose a customized algorithm for category-level articulated object pose estimation, aiming to advance the tracking performance on the SE(3) manifolds.
    \item We propose a weighted point pair algorithm to predict SE(3)-invariant parameters and estimate pose increments in the associated Lie algebra $\mathfrak{se}(3)$. To respect the kinematic constraints of articulated objects, an axis-based part-level refinement is further introduced for improved accuracy and consistency.
    \item Extensive experiments on both point cloud and RGB-D datasets demonstrate the superior performance and efficiency of our method. Moreover, real-world evaluations confirm its strong generalization capability across diverse and challenging environments.
\end{itemize}

\section{RELATED WORK} \label{sec:RELATED_WORK}
\noindent $\bigstar$ \textbf{Category-Level Object Pose Estimation.}
Category-level object pose estimation aims to predict the 6D pose (rotation and translation) and scale of previously unseen instances within known categories~\cite{zhang2025u,liu2022toward,zhang2025r}. Early works like NOCS~\cite{wang2019normalized} proposed normalized coordinate spaces to enhance generalization. DualPoseNet~\cite{lin2021dualposenet} introduced dual decoders for joint implicit and explicit modeling. Recent methods~\cite{lin2022keypoint,irshad2022shapo,you2022ukpgan,fernandez2020unsupervised} leverage shape priors and keypoint voting to handle intra-class variation, but often rely on extensive real or synthetic training data~\cite{liu2022akb,you2022go}, limiting scalability in complex real-world environments~\cite{di2022gpv,zhang2024generative}. While self-supervised approaches~\cite{deng2020self,manhardt2020cps++,chen2024bridging} reduce annotation dependency, they still suffer from domain shifts~\cite{sankaranarayanan2018learning} and background assumptions~\cite{hurlburt2011presuppositions}.

\noindent $\bigstar$ \textbf{Articulated Object Pose Tracking.}
Articulated pose tracking extends the estimation task to dynamic, multi-frame settings~\cite{zhang2024vocapter,zhang2021keypoint}. Instance-level methods such as BundleTrack~\cite{wen2021bundletrack} and PA-Pose~\cite{liu2024pa} rely on segmentation and alignment optimization, while keypoint-based methods~\cite{maji2022yolo,li2021tokenpose,prakhya2015sparse,jau2020deep} struggle with occlusion and articulation variability. RPMArt~\cite{wang2024rpmart} combines deep segmentation with robust features for category-level tracking, but high computational cost limits real-time applicability. A major drawback of existing methods lies in their frame-by-frame processing, which ignores inter-frame motion continuity and structural consistency in SE(3).

\begin{figure*}
    \centering
    \includegraphics[width=0.9\linewidth]{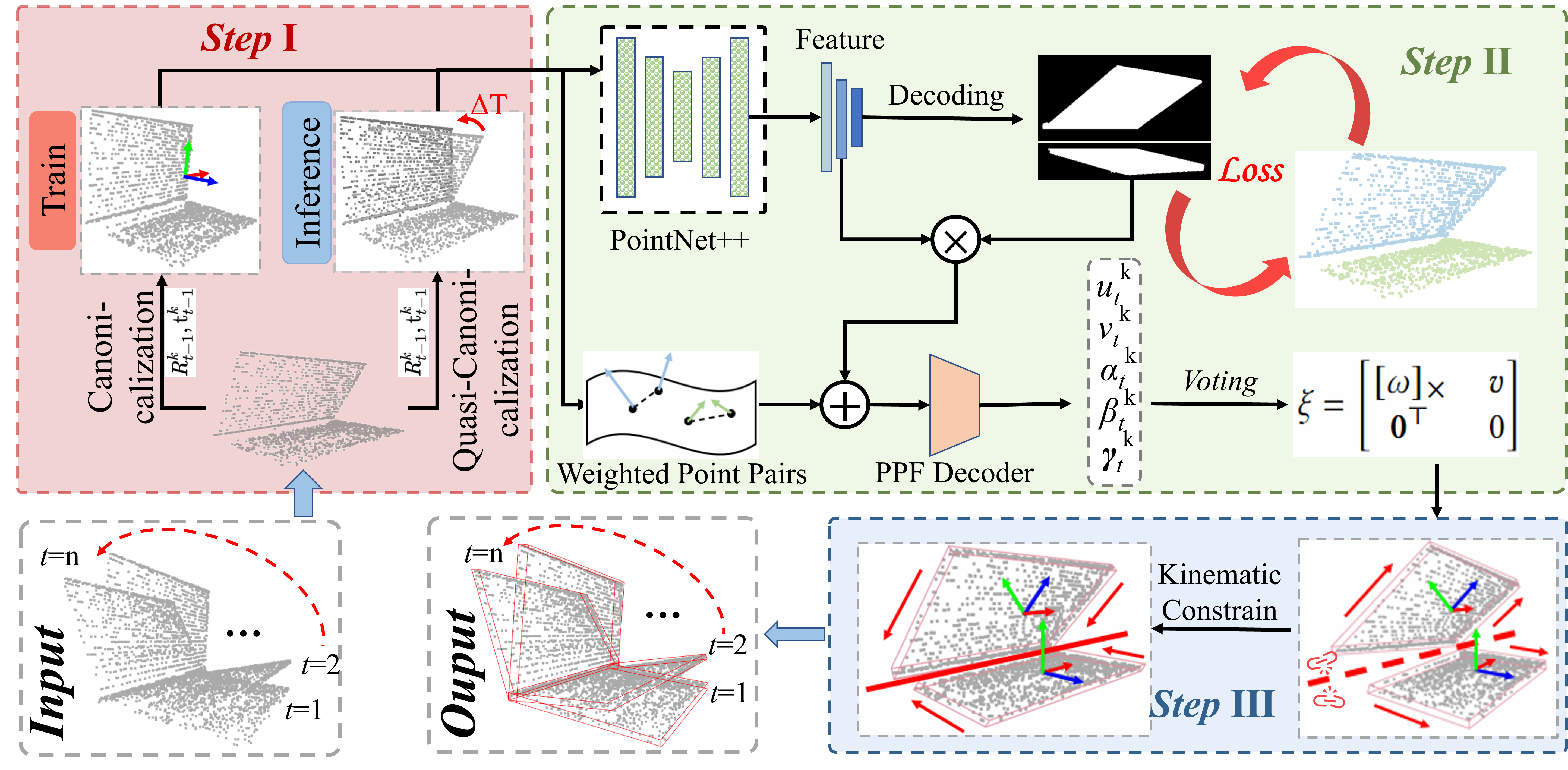}
    \caption{The Overview of Our PPF-Tracker.
    }
    \label{fig:overview}
\end{figure*}

\section{Problem Statement}

This work addresses the category-level articulated object pose tracking problem on the $\mathrm{SE}(3)$ manifolds. Given a sequence of frames $\{P_t\}_{t \geq 0}$ and the initial pose $T_0^k$ as input, our goal is to estimate the per-part pose and scale for each frame. Assuming the articulated object consists of $K$ rigid parts and $J$ joints, we use the superscript $k$ to denote the $k$-th part and $j$ for the $j$-th joint. The output of our method is the part-wise pose $T_t^k \in \mathrm{SE}(3)$ and the corresponding scale $s_t^k \in \mathbb{R}^3$. The transformation matrix $T_t^k$ is defined as:
\begin{equation}
T_t^k = 
\begin{bmatrix}
R_t^k & \mathbf{t}_t^k \\
\mathbf{0}^T & 1
\end{bmatrix}, \quad R_t^k \in \mathrm{SO}(3),~\mathbf{t}_t^k \in \mathbb{R}^3
\end{equation}
where $R_t^k$ represents the rotation and $\mathbf{t}_t^k$ denotes the translation of the $k$-th part at time $t$.

The tracking task is formulated as an inter-frame pose increment estimation problem. Specifically, given the pose in the previous frame, denoted as $T_{t-1}^k$, the pose in the current frame $T_t^k$ can be expressed as:
\begin{equation} \label{eq:T_t^k}
    T_t^k = \Delta T_t^k \cdot T_{t-1}^k
\end{equation}
\noindent where $T \in \mathrm{SE}(3)$. $\Delta T_t^k$ represents the pose increment from the previous frame to the current frame. Furthermore, we adopt Lie algebra to model the pose, which allows us to reformulate Eq.~\ref{eq:T_t^k} as:
\begin{equation}
    T_t^k = \exp(\xi_t^k) = \exp(\Delta \xi_t^k + \xi_{t-1}^k)
\end{equation}

\noindent Here, $\xi \in \mathfrak{se}(3)$ represents the pose in the Lie algebra. The exponential map $\exp(\cdot)$ projects elements from the Lie algebra to the Lie group. As such, both $\xi$ and $T$ provide valid representations of a 6D pose.

\section{METHOD} \label{sec:METHOD}
\subsection{Quasi-Canonicalization} \label{sec:Canonicalization}
\begin{figure}
    \centering
    \includegraphics[width=0.96\linewidth]{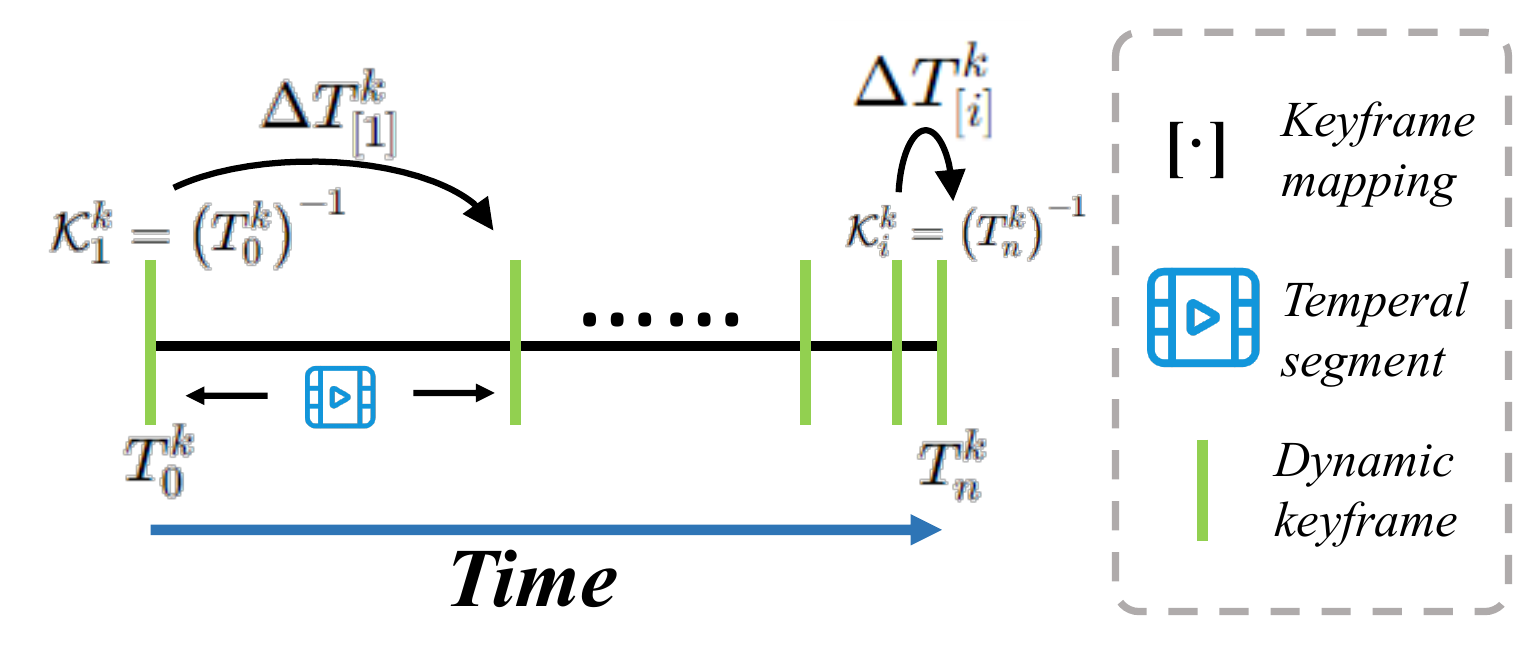}
    \caption{Illustration of Temperal Segment and Dynamic Keyframe. The symbols $\mathcal{K}_i^k$ and $T_n^k$ are the poses of the $i$-th keyframe and the $n$-th frame, respectively. The subscript symbol $[\cdot]$ denotes the mapping of the keyframe to the frame stream. For instance, $[1]=0$ and $[i]=n$.}
    \label{fig:initframe}
\end{figure}

To leverage temporal priors across consecutive frames, we introduce a \textbf{Quasi-Canonicalization} strategy. Specifically, the point cloud sequence is divided into temporal segments, each bounded by two consecutively selected keyframes, as illustrated in Fig.~\ref{fig:initframe}. For the $i$-th segment, we designate its first frame as the keyframe and define the associated transformation as $\mathcal{K}_i^k = (T_n^k)^{-1}$, where $n$ denotes the keyframe’s original position in the sequence. This definition establishes a one-to-one mapping between the keyframe index $i$ and its temporal index $n$.

\begin{figure}
    \centering
    \includegraphics[width=0.8\linewidth]{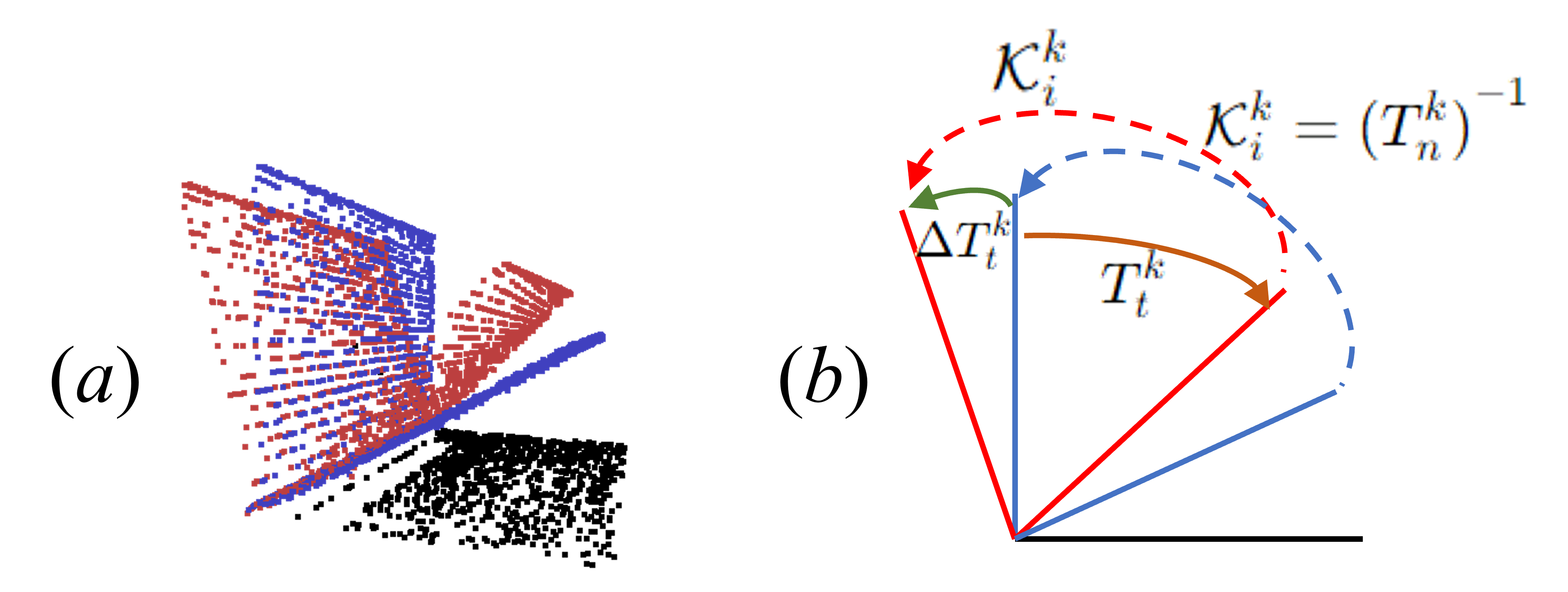}
    \caption{The Illustration of Quasi-Canonicalization within \textit{Temporal Segment}. For clearer expression, we abstract (a) as (b). The \textbf{blue} part represents the keyframe transformation (canonicalization). The \textbf{red} part depicts the quasi-canonicalization of the $t$-th frame where the frame follows its associated keyframe's transformation.}
    \label{fig:incresement}
\end{figure}

We formally define the operation of applying the inverse transformation $(T_t^k)^{-1}$ to the $t$-th frame as \textit{Canonicalization}. In parallel, the application of $\mathcal{K}_i^k$ to all frames within the $i$-th segment constitutes our \textit{Quasi-Canonicalization} process. As shown in Fig.~\ref{fig:incresement}, the resulting point cloud in the quasi-canonical space, denoted as $\bar{\mathcal{P}}_t^k$, is formulated as:
\begin{equation}
    \bar{\mathcal{P}}_t^{k}=\Delta T_t^k \mathcal{P}_c^k
\end{equation}

\noindent where $\mathcal{P}_c^k$ is the point cloud in the canonical space, $T_t^k$ is the transformation of the \(t\)-th frame, representing its absolute pose defined in the camera space, and $\mathcal{K}_i^k$ is the pose of the $i$-th keyframe.

The resulting quasi-canonical point cloud $\bar{\mathcal{P}}_t^k$ is then used as input for predicting the relative transformation $\Delta T_t^k$. The corresponding ground truth is defined as:
\begin{equation}
    \Delta T_t^{k(*)}=T_t^k (\mathcal{K}_i^k)^{-1}
    \label{Eq:1}
\end{equation}
\noindent where $(*)$ represents the ground truth. It is worth noting that our method does not directly regress $\Delta T_t^k$. Instead, a more accurate prediction strategy is employed, as detailed in Sec.~
\ref{sec:4.2}.

Recall Eq.~\ref{Eq:1}, $T_t^k$ is with the $i$-th keyframe as the beginning. We extend this formulation to the initial frame with absolute pose $T_{0}^k$. In pose tracking task, the first absolute pose is always $T_0^k$ as shown in Fig.~\ref{fig:initframe}. So the pose $T_t^k$ can be formulated as: 
\begin{equation}
    T_t^k=\Delta T_{[i]}^k \Delta T_{[i-1]}^k \cdots \Delta T_{[1]}^k T_{0}^k=(\prod_i \Delta T_{[i]}^k)  T_{0}^k
    \label{Eq:4}
\end{equation}
Despite the keyframe method being able to mitigate cumulative errors, we find that the fixed keyframe strategy lacks flexibility, which leads to inaccuracy, as seen in Section~\ref{sec:ablationstudy}. In this work, we propose a \textbf{Dynamic Keyframe Selection (DKS)} strategy. Our DKS strategy intelligently incorporates an energy function, where the keyframe is updated whenever the energy value falls below a predefined threshold $\phi=0.01$. Specifically, we compute both the chamfer distance $D_C$ and the hausdorff distance $D_H$ between the predicted point cloud $\hat{\mathcal{P}}$ and the actually observed $\mathcal{P}$. Then, the energy function can be formulated as:
\begin{equation}
    \mathfrak{E}_t = \frac{1}{|\mathcal{P}|}(D_C + D_H)
\end{equation}
\noindent where $|\mathcal{P}|$ is the the number of points in $\mathcal{P}$. If $\mathfrak{E}_t < \phi$, it indicates a high similarity between the predicted frame and the observed frame, demonstrating the frame's reliability. Therefore, it is selected as the keyframe for the next \textit{temporal segment}. 

\subsection{SE(3)-Invariance based Increment Learning} \label{sec:4.2}
\begin{figure}
    \centering
    \includegraphics[width=0.8\linewidth]{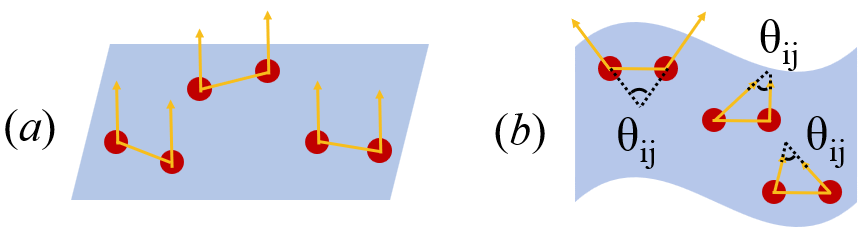}
    \caption{Traditional (a) and Weighted (b) Point Pairs.}
    \label{fig:weighted ppf}
\end{figure}

To ensure robust and accurate incremental estimation, we do not directly predict $T_t^k$. Instead, we adopt a PPF voting strategy to infer SE(3)-invariant parameters, from which a Lie algebra representation is obtained through voting, and the final transformation $T_t^k$ is then derived via the exponential map. The overall process of our method consists of two key steps: \textbf{SE(3)-invariant parameters prediction} and \textbf{Lie algebra transformation}, detailed as follows:

\textbf{(1) SE(3)-invariant parameters prediction}:  
The PPF describes 3D shape characteristics by computing the relative geometric relationships between neighboring point pairs in a local coordinate system, thus inherently possessing SE(3) invariance. However, conventional point pair descriptors assign equal importance to all point pairs (Fig.~\ref{fig:weighted ppf} (a))~\cite{you2022cppf}. Intuitively, different point pairs play distinct roles in describing the three-dimensional features of objects (Fig.~\ref{fig:weighted ppf} (b)). Therefore, we use the \textbf{weighted PPF}.

Specifically, given the observed point cloud $\mathcal{P}_t$, we uniformly sample $N$ point pairs $(\mathbf{p}_i, \mathbf{p}_j) \in (\mathcal{P}_t,\mathcal{P}_t)$, where all the sampled point $p \in \mathcal{P}_t$. For each pair, we compute an SE(3)-invariant feature vector \(\mathcal{F}_{ij}\) using the positions ($\mathbf{p}_i, \mathbf{p}_j$) and the surface normals ($\mathbf{n}_i, \mathbf{n}_j$) of the point pairs. This feature ensures invariance to rigid transformations, enabling generalization across unseen object instances~\cite{lin2022pose}. The weight \(v_{ij}\) for each point pair \((\mathbf{p}_{i},\mathbf{p}_{j})\) is determined based on the angle \(\theta_{ij}\) between their surface normals:
\begin{equation}
    v_{ij}=1-\lambda|\cos\theta_{ij}|
    \label{eq:weighted}
\end{equation}
\noindent where \(\lambda\ = 0.5\) is a tunable parameter. Eq.~\ref{eq:weighted} assigns lower weights to point pairs with near-parallel normals (where \(\theta_{ij}\approx 0^{\circ}\) or \(180^{\circ}\)) and higher weights to pairs with near-perpendicular normals (where \(\theta_{ij}\approx 90^{\circ}\)).

Then PointNet++~\cite{qi2017pointnet++} is used to process $\mathcal{F}_{ij}$ to predict per part SE(3)-invariant parameters $(\mu_t^{ k},\nu_t^{ k},\alpha_t^{ k},\beta_t^{ k},\gamma_t^{ k})$ where $(\mu_t^{ k},\nu_t^{ k})$, $(\alpha_t^{ k},\beta_t^{ k})$ and $\gamma_t^{ k}$ related to the translation, rotation and scale parameters respectively. We denote the object center as $\mathbf{o}$, up orientation as $\overrightarrow{\mathbf{e}}_1$, right orientation as $\overrightarrow{\mathbf{e}}_2$ and unit vector $   \overrightarrow{\mathbf{d}}=\frac{\overrightarrow{\mathbf{p}_i\mathbf{p}_j}}{\|\overrightarrow{\mathbf{p}_i\mathbf{p}_j}\|_2}$. These SE(3)-invariant parameters can be formulated as:
\begin{equation}
    \left\{
    \begin{aligned}
    \mu_t^{ k} &= \overrightarrow{\mathbf{p}_i\mathbf{o}} \cdot \overrightarrow{\mathbf{d}}, \\
    \nu_t^{ k} &= \|\overrightarrow{\mathbf{p}_i\mathbf{o}}-\mu_t^{ k} \overrightarrow{\mathbf{d}})\|_2, \\
    \alpha_t^{ k} &= \overrightarrow{\mathbf{e}}_1 \cdot \overrightarrow{\mathbf{d}}, \\
    \beta_t^{ k} &= \overrightarrow{\mathbf{e}}_2 \cdot \overrightarrow{\mathbf{d}}.
    \end{aligned}
    \right.
\end{equation}

\begin{figure}
    \centering
    \includegraphics[width=0.7\linewidth]{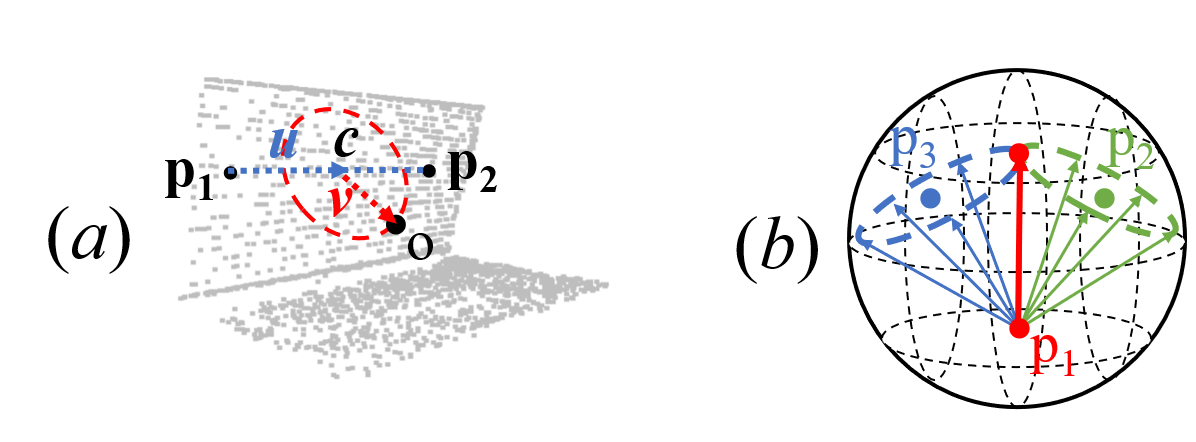}
    \caption{Illustration of Voting Scheme. (a) For orientation voting scheme, the center of each point pair is determined by a circle which is divided into bins. (b) For orientation voting scheme, a Fibonacci sphere with uniformly distributed bins is used for voting accumulation.}
    \label{fig:voting}
\end{figure}

As depicted in Fig.~\ref{fig:voting}, the per-part object center is constrained to lie on the circle defined by the translation parameters \((\mu_t^{k}, \nu_t^{k})\), where \(\mu_t^k\) determines the center \(c\) of the circle, and \(\nu_t^k\) defines its radius. The orientation of the per-part object is determined by conics, which are derived from votes generated by multiple point pairs.

The remaining pose parameter $\gamma_t^{ k}$ represents the scaling factor which can be predicted directly. 

\textbf{(2) Lie algebra transformation}: 
To avoid numerical divergence or loss of orthogonality that may arise from directly voting for $\Delta T_t^k$, we first vote for elements in the Lie algebra tangle space where \(\Delta\xi \in \mathfrak{se}(3)\), and then the multiplication of transformation matrices can be converted into the addition of their corresponding Lie algebras. 

With the SE(3)-invariant parameters, we can transform them into Lie algebra according to ~\cite{eade2013lie}. We define $\xi$ as the Lie algebra element by constructing a $4 \times 4$ matrix:

\begin{equation}
\xi = \begin{bmatrix}
[\omega]_\times & v \\
\mathbf{0}^\top & 0
\end{bmatrix} \in \mathfrak{se}(3),
\end{equation}
where $\omega \in \mathbb{R}^3$ represents rotation parameter and $v \in \mathbb{R}^3$ represents translation parameter. The symbol $[\cdot]_{\times}$ means the skew-symmetric matrix.

Utilizing Lie Algebra increments for pose tracking inherently offers more stable performance~\cite{li2021leveraging}. Recall Eq.~\ref{Eq:4}, we can re-formulate it as:
\begin{equation}
    \xi_t^{ k}=\Sigma_i{\Delta \xi_{[i]}^k} + \xi_{0}^k 
\end{equation}
\noindent where $\xi^k$ is the Lie Algebra of $T^k$, and $\Delta\xi^k$ is the Lie Algebra of $\Delta T^k$. Through the exponential map $T_t^k = exp(\xi_t^k)$, this linearized variation can be remapped to the SE(3) manifolds. This mapping inherently satisfies the orthogonality constraint of the rotation matrix, effectively preventing the output of non-orthogonal matrices. In our model, the angular velocity vector \(w \in \mathbb{R}^3\) is free from the gimbal lock problem and maintains a one-to-one correspondence with the rotation matrix within the range \(\|w\| < \pi\), thereby ensuring accuracy and consistency in pose estimation. 

\subsection{Optimization via Kinematic Constraints} \label{sec:4.3}
Since our SE(3)-invariant framework models each part as an independent rigid body, this assumption may result in discontinuities across kinematic chains, leading to physically inconsistent motion. To mitigate this issue, we introduce a \textbf{Kinematic-Constrained Optimization} strategy that enforces rigid coupling between connected parts along their articulated axes.

Specifically, our method begins with coarse per-part pose predictions obtained through PPF voting, which provides a robust initialization for subsequent refinement. We then propose a comprehensive energy function that simultaneously minimizes geometric alignment errors while enforcing kinematic joint constraints. For per-part of the observed point cloud \(\mathcal{P}_t^k\) and the canonical point cloud $\mathcal{P}_c^k$, the geometric alignment term $\mathfrak{E}_{geo}$ is defined as:
\begin{equation}
\mathfrak{E}_{geo} =  \sum_{k=1}^{K} \left\| \frac{1}{|N^k|}(T^k)^{-1} \mathcal{P}^k - \mathcal{P}_c^k) \right\|^2,
\end{equation}

For the kinematic term $\mathfrak{E}_{kin}$, we take the $j$-th axis point $q^j$ as an example, homogeneous normalization is applied to the axis point $(q^j)'=[q^j, 1]$. The kinematic term can be formulated as:
\begin{equation}
\mathfrak{E}_{kin} = \sum_{j=1}^{J-1} \left\| T^j (q_j')^T -  T^{j+1}(q_j')^T) \right\|^2
\end{equation}
\noindent where $T^j$ and $T^{j+1}$ represent the transformations of two parts connected by the same axial joint.
Finally, our comprehensive energy function is given by  $\mathfrak{E}_{comp} = \mathfrak{E}_{geo} + \mathfrak{E}_{kin}$.

With the comprehensive energy function, we can optimize $\hat{T}_t^k$ as $(\hat{T}_t^k)_{optim}$. Our optimization reduces system degrees of freedom and eliminates pose ambiguity. By incorporating kinematic constraints that reflect the geometric configuration of articulated structures, we restrict the motion space, thereby enhancing physical plausibility and ensuring more consistent pose tracking.

$\bigstar$ The overall articulation tracking procedure for the frame stream is summarized in Algorithm ~\ref{alg:1}.

\begin{table*}[tbh]
    \small
    \centering
    \resizebox{0.9\linewidth}{!}{
    \begin{tabular}{c l ccc c}
    \toprule
    \multirow{2}{*}{\textbf{Category}} & \multirow{2}{*}{\textbf{Method}} & \multicolumn{3}{c}{\textbf{Per-part 6D Pose}} & \multirow{2}{*}{\textbf{Inference Time (s)} $\downarrow$} \\
    \cline{3-5}
    & & Rotation Error ($^{\circ}$) $\downarrow$ & Translation Error ($m$) $\downarrow$ & 3D IOU (\%) $\uparrow$ \\
    \hline
    \multirow{5}{*}{Laptop} & A-NCSH \cite{li2020category} & 8.5, 9.2 & 0.084, 0.103 & 40.8, 28.3 & 1.67 \\
    & CAPTRA \cite{weng2021captra} & 5.9, 5.3 & 0.080, 0.063 & \textbf{70.1}, 43.5 & 0.10 \\
    & ContactArt \cite{zhu2024contactart} & 8.8, 9.2 & 0.101, 0.154 & 50.2, 36.5 & 0.71 \\
    & GAPS~\cite{yu2025generalizable} & 8.7, 8.9 & 0.092, 0.096 & 54.3, 39.3 & 0.34 \\
    & \textbf{PPF-Tracker} (Ours) & \textbf{3.7}, \textbf{4.5} & \textbf{0.043}, \textbf{0.055} & 68.5, \textbf{49.6} & \textbf{0.07} \\
    \hline
    
    \multirow{5}{*}{Eyeglasses} & A-NCSH \cite{li2020category} & 7.6, 24.8, 26.6 & 0.079, 0.324, 0.319 & 40.5, 29.3, 28.4 & 2.59 \\
    & CAPTRA \cite{weng2021captra} & 4.5, 12.6, 13.1 & 0.054, 0.097, 0.084 & 53.1, 41.2, 39.8 & 0.14 \\
    & ContactArt \cite{zhu2024contactart} & 11.3, 16.5, 13.2 & 0.134, 0.158, 0.137 & 47.3, 47.4, 43.2 & 1.00 \\
    & GAPS~\cite{yu2025generalizable} & 8.5, 9.3, 9.6 & 0.105, 0.123, 0.118 & 48.2, 36.7, 35.6 & 0.84 \\
    & \textbf{PPF-Tracker} (Ours) & \textbf{2.5}, \textbf{3.6}, \textbf{3.8} & \textbf{0.031}, \textbf{0.038}, \textbf{0.041} & \textbf{58.6}, \textbf{55.7}, \textbf{59.2} & \textbf{0.12} \\
    \hline
    
    \multirow{5}{*}{Dishwasher} & A-NCSH \cite{li2020category} & 5.0, 5.7 & 0.074, 0.119 & 64.5, 43.8 & 1.70 \\
    & CAPTRA \cite{weng2021captra} & 4.6, 5.4 & 0.055, 0.089 & 85.3, 61.2 & 0.11 \\
    & ContactArt \cite{zhu2024contactart} & 7.8, 6.9 & 0.106, 0.145 & 78.3, 65.2 & 0.67 \\
    & GAPS~\cite{yu2025generalizable} & 6.2, 7.0 & 0.126, 0.207 & 80.3, 54.3 & 0.36 \\
    & \textbf{PPF-Tracker} (Ours) & \textbf{3.2}, \textbf{3.4} & \textbf{ 0.038}, \textbf{0.045} & \textbf{87.2}, \textbf{76.1} & \textbf{0.06} \\
    \hline
    
    \multirow{5}{*}{Scissors} & A-NCSH \cite{li2020category} & 5.0, 5.7 & 0.041, 0.057 & 32.3, 32.8 & 1.21 \\
    & CAPTRA \cite{weng2021captra} & 4.1, 4.7 & 0.032, 0.039 & 43.2, 42.8 & 0.12 \\
    & ContactArt \cite{zhu2024contactart} & 6.8, 7.5 & 0.085, 0.067 & 39.6, 38.0 & 0.5 \\
    & GAPS~\cite{yu2025generalizable} & 6.1, 6.6 & 0.055, 0.069 & 41.3, 40.2 & 0.29 \\
    & \textbf{PPF-Tracker} (Ours) & \textbf{3.4}, \textbf{4.1} & \textbf{0.027}, \textbf{0.033} & \textbf{45.2}, \textbf{44.6} & \textbf{0.09} \\
    \hline
    
    \multirow{5}{*}{Drawer} & A-NCSH \cite{li2020category} & 8.6, 9.8, 11.5, 8.5 & \textbf{0.088}, 0.255, 0.257, 0.175 & 66.5, 62.3, 58.6, 61.3 & 3.64 \\
    & CAPTRA \cite{weng2021captra} & \textbf{4.8}, 6.5, 6.3, 6.0 & 0.112, 0.185, 0.177, 0.156 & 89.5, 80.1, 76.7, 78.2 & 0.25 \\
    & ContactArt \cite{zhu2024contactart} & 6.5, 7.8, 7.6, 8.1 & 0.132, 0.172, 0.188, 0.207 & 71.2, 70.6, 68.3, 69.4 & 1.00 \\
    & GAPS~\cite{yu2025generalizable} & 6.5, 6.5, 6.5, 6.5 & 0.168, 0.242, 0.243, 0.239 & 86.3, 76.5, 75.6, 77.6 & 0.62 \\
    & \textbf{PPF-Tracker} (Ours) & 5.2, \textbf{5.2}, \textbf{5.2}, \textbf{5.2} & 0.095, \textbf{0.172}, \textbf{0.162}, \textbf{0.153} & \textbf{90.2}, \textbf{85.1}, \textbf{78.6}, \textbf{80.9} & \textbf{0.16} \\
    \bottomrule
    \end{tabular}}
    \caption{Comparison with SOTA Methods on PM-Videos Dataset.}
    \label{tab:artimage}
\end{table*}
\begin{figure*}[t!]
    \centering
    \includegraphics[width=0.88\linewidth]{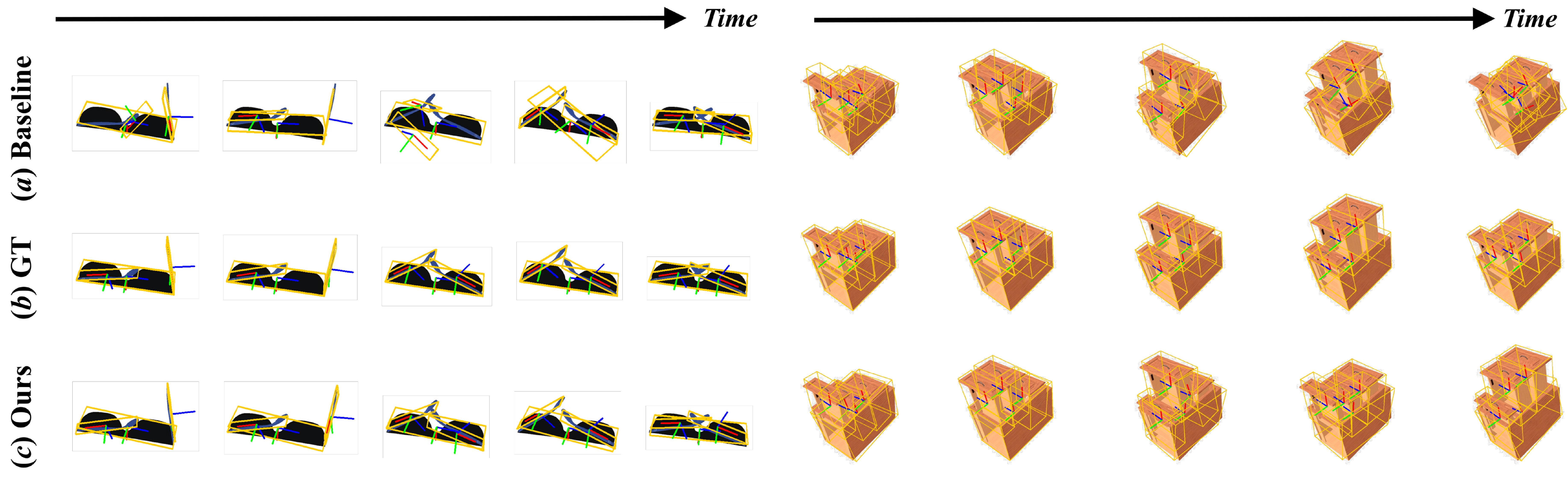}
    \caption{Qualitative Results on PM-Videos Dataset. }
    \label{fig:ArtImage}
\end{figure*}

\begin{algorithm}[ht]
\begin{algorithmic}[1]
    \STATE \textbf{Input}: The frame stream $\{\mathcal{P}_t^{k}\}_{t \geq 0}$ and initial pose $\xi_0^k$\\
    
    \STATE \textbf{Output}: Per-part 6D pose $(\hat{T}_t^{k})_{optim}$ and scale $\hat{s}_{t}^{k}$ for all the $t>0$ frames.
    \STATE Initialize keyframe as $\mathcal{P}_0^{k}$.
    \FOR{\textbf{each} rigid part $\mathcal{P}_t^{k} \in \{\mathcal{P}_t^{k}\}_{t>0}$}
       
        \STATE Sample $N$ weighted point pairs $(\mathbf{p}_i, \mathbf{p}_j) \in (\mathcal{P}_t,\mathcal{P}_t)$.
       
        \STATE Predict the SE(3)-invariant parameters.      
        \STATE Vote for the Lie algebra element ${\Delta \hat{\xi}}_{t}^{k}$.
        \STATE Accumulate increments $\hat{\xi}_t^{ k}=\hat{\xi}_{t-1}^k + {\Delta \hat{\xi}}_{t}^{k}$.
        
        \STATE Compute coarse per-part pose $\hat{T}_t^k$ and scale $\hat{\textbf{s}}_{t}^{k}$
        \STATE Kinematic constraints: $\hat{T}_t^k  \rightarrow (\hat{T}_t^k)_{optim}$.     
        \IF{$\mathfrak{E}_{t} < \phi$}
            \STATE Update keyframe.
        \ENDIF
        
    \ENDFOR
    \caption{PPF-Tracker: Category-level Articulated Object Pose Tracking on SE(3) Manifolds.}
    \label{alg:1}
\end{algorithmic}
\end{algorithm}

\subsection{Loss Functions}\label{4.4}
We adopt the KL divergence as the loss function to quantify the discrepancy between probability distributions, with particular emphasis on deviations in low-probability regions. By applying the KL divergence to point-wise feature voting outcomes, we promote both efficient network convergence and computational efficiency. For scale prediction, we employ the MSE loss to quantify the difference between the predicted and ground-truth scales. For mask learning, we adopt the BCE loss to supervise the predicted binary masks against the ground truth. 

The detailed computation of the loss functions is provided in the Supplementary Material. Here, the total loss is formulated as a weighted sum of the aforementioned components:
\begin{equation}
\mathcal{L}_{\text{total}}=\lambda_{1}\mathcal{L}_{\text{trans}}+\lambda_{2}\mathcal{L}_{\text{orient}}+\lambda_{3}\mathcal{L}_{\text{scale}}+\lambda_4\mathcal{L}_{mask}
\end{equation}
\noindent where \(\lambda_{1}=0.3\), \(\lambda_{2}=0.3\), \(\lambda_{3}=0.2\), and \(\lambda_{4}=0.2\) are hyperparameters that balance the contributions of each loss term. This comprehensive loss function ensures that the model effectively learns the geometric relationships and SE(3)-invariant features, leading to robust and accurate pose tracking results.

\section{EXPERIMENT} \label{sec:EXPERIMENT}
\subsection{Experimental Setup}
\noindent\textbf{Implementation.} During data pre-processing, input point cloud is downsampled to 3,072 points, and objects in RGB-D images are cropped and projected into the point cloud to serve as network inputs. The initial learning rate is set to 0.001, decreasing by a factor of 0.1 every 10 epochs. The number of total training epochs is 200. All the experiments are implemented on an NVIDIA GeForce RTX 4090 GPU with 24GB of memory.

\noindent \textbf{Datasets and metrics.} Following the setup in~\cite{weng2021captra}, we construct a synthetic dataset for category-level articulated object pose tracking based on PartNet-Mobility~\cite{xiang2020sapien}. As the original data generation scripts and configuration details were not released, direct reproduction was not feasible. To ensure scientific rigor and comparability, we rebuild the synthetic tracking dataset using a similar strategy.
Concretely, we generate a synthetic tracking dataset named PM-Videos from PartNet-Mobility~\cite{xiang2020sapien}, semi-synthetic tracking dataset named ReArt-Videos from ReArt-48 repository~\cite{liu2022akb}, and Real-world Scenario tracking dataset named RobotArm-Videos from RobotArm dataset. 
To comprehensively evaluate the performance of PPF-Tracker, multiple metrics are adopted: degree error ($^\circ$) for rotation, distance error ($m$) for translation, 3D IOU (\%) for scale, and tracking speed ($s$) for real-time performance. We also measure cumulative tracking error across entire video to better quantify long-term tracking ability.

\subsection{Comparison with the SOTA Methods}
In this section, we evaluate the proposed PPF-Tracker on the synthetic articulated object dataset PM-Videos, which are generated from PartNet-Mobility~\cite{xiang2020sapien}, with quantitative results summarized in Table~\ref{tab:artimage}. Compared to previous methods such as A-NCSH~\cite{li2020category} and ContactArt~\cite{zhu2024contactart}, PPF-Tracker consistently achieves superior tracking performance across all object parts, as evidenced by significantly lower rotation and translation errors.

Specifically, for the \textit{Eyeglasses} category, our method achieves an average rotation error of \textbf{3.3}$^\circ$ and a translation error of 0.036$m$, representing a relative reduction of approximately \textbf{60}\% compared to the second-best method, GAPS. Additionally, the 3D IoU shows a notable improvement, with an average increase of \textbf{17.6}\%. In terms of runtime performance, PPF-Tracker demonstrates strong real-time capabilities, with the inference time \textbf{0.12}$s$ per frame.

Qualitative tracking results are illustrated in Fig.~\ref{fig:ArtImage}. We hypothesize that PPF-Tracker’s superior performance arises from the combination of voting-based strategies and Lie algebra-based transformation. Together, these provide a robust trade-off between tracking accuracy and efficiency.

\subsection{Ablation Study} \label{sec:ablationstudy}

\noindent\textbf{Kinematic Constraints.}
To evaluate the effectiveness of the proposed Kinematic Constraints, we conduct ablation experiments, with results summarized in Table~\ref{tab:ablation1}. The application of kinematic constraints results in substantial error reduction—rotation errors decrease by \textbf{5.4$^{\circ}$} and \textbf{5.7$^{\circ}$}, while translation errors are reduced by \textbf{0.071$m$} and \textbf{0.077$m$}, amounting to over \textbf{50\%} improvement compared to the unconstrained setting. \textit{These results underscore the necessity of incorporating kinematic constraints when performing part-level pose tracking, as they effectively enforce structural consistency across articulated components.}

\noindent\textbf{Keyframe Seletion.}
To validate the effectiveness of the dynamic keyframe strategy, we designed the following ablation experiment to evaluate three different settings: 1) No Keyframe, 2) Fixed Keyframe, and 3) Dynamic Keyframe. The experimental results are shown in Table~\ref{tab:ablation1}, and all experiments were conducted on the same dataset. The errors are largest when no keyframes are used. \textit{This indicates that without keyframes, accumulated errors quickly grow, leading to a significant decline in pose tracking accuracy. The fixed keyframe strategy somewhat alleviates the issue of accumulated errors but still underperforms compared to the dynamic keyframe approach.}

\begin{table}[t!]
\begin{center}
\resizebox{0.95\linewidth}{!}{
\begin{tabular}{cccc}
\toprule
& \multirow{2}{*}{\textbf{Kinematic Constraints}} & \multicolumn{2}{c}{\textbf{Per-part 6D Pose}}\\
\cmidrule(lr){3-4}
&   & Rotation Error ($^\circ$) & Translation Error ($m$) \\
\midrule
I &  $\times$ & 8.6, 9.1 & 0.109, 0.122\\
II~(Ours)  &  \checkmark  & \textbf{3.2}, \textbf{3.4} & \textbf{0.038}, \textbf{0.045}\\
\midrule
& \multirow{2}{*}{\textbf{Keyframe Selection}} & \multicolumn{2}{c}{\textbf{Per-part 6D Pose}}\\
\cmidrule(lr){3-4}
&   & Rotation Error ($^\circ$) & Translation Error ($m$) \\
\midrule
III & No Keyframe & 13.6, 15.0 & 0.125, 0.153\\
IV & Fixed Keyframe & 5.3, 6.2 & 0.061, 0.087\\
V~(Ours) & Dynamic Keyframe & \textbf{3.2}, \textbf{3.4} & \textbf{0.038}, \textbf{0.045}\\
\bottomrule
\end{tabular}}
\caption{Ablation Experiments with the PPF-Tracker.}
\label{tab:ablation1}
\end{center}   
\end{table}

\subsection{Generalization Capacity}
\begin{figure}[t!]
    \centering
    \includegraphics[width=\linewidth]{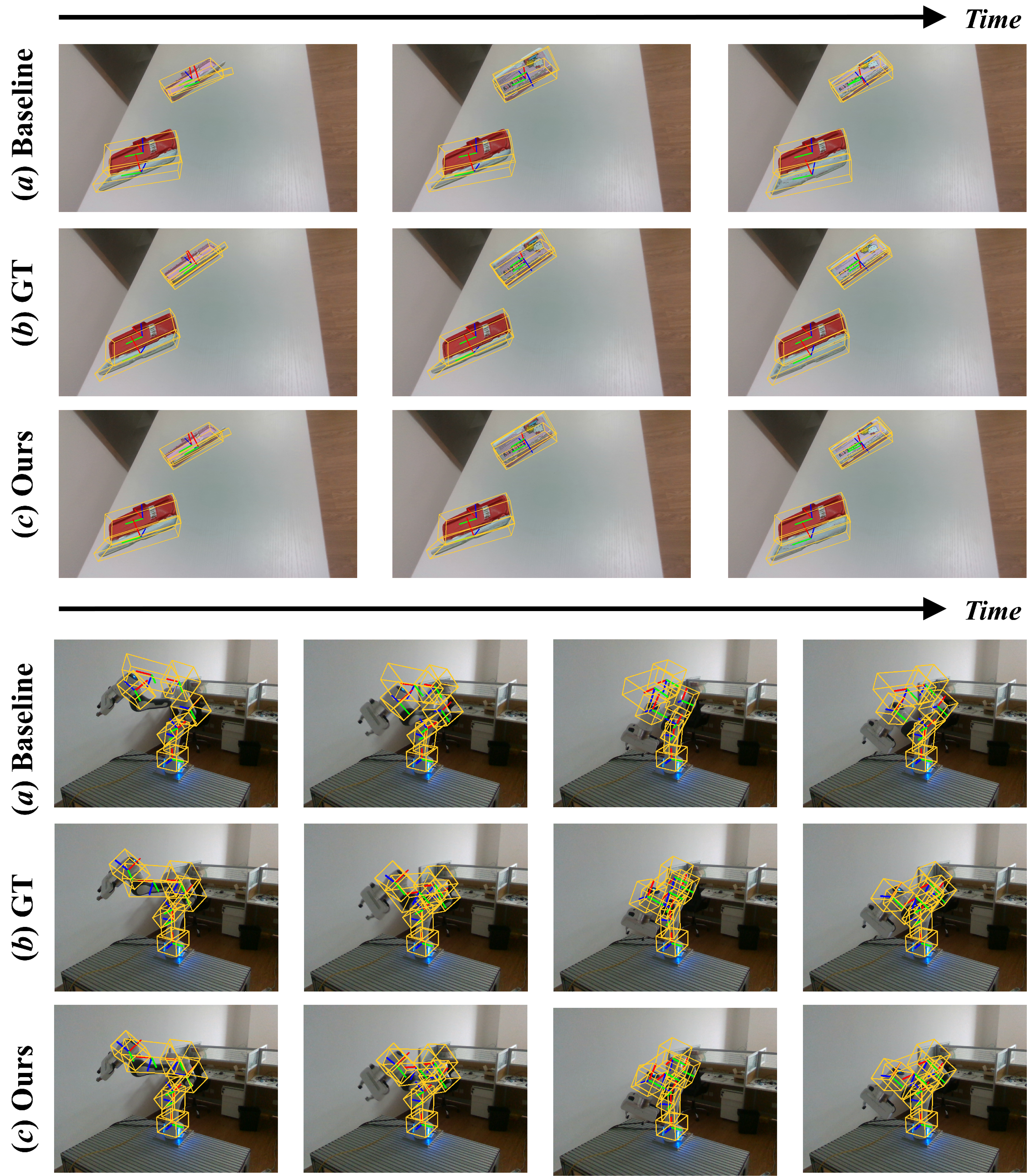}
    \caption{Qualitative Results on ReArt-Videos (Top) and RobotArm-Videos (Bottom).}
    \label{fig:ReArtMix-Franka}
\end{figure}

\textbf{Experiments on Semi-Synthetic Scenarios.}  We evaluate the effect of our PPF-Tracker on the semi-synthetic dataset ReArt-Videos. Results are shown in Fig.~\ref{fig:ReArtMix-Franka} (Top) and Table~\ref{tab:reart_exp}. The tracking results show that our method can perform well in the semi-synthetic scenarios.

\noindent \textbf{Experiments on Real-world Scenarios.} To investigate the tracking performance in real-world scenarios, we also evaluate the proposed PPF-Tracker on the 7-part RobotArm-Videos. Fig.~\ref{fig:ReArtMix-Franka} (Bottom) and Table~\ref{tab:robotarm_exp} show the qualitative and quantitative results, respectively. It is evident that
our approach achieves acceptable 6D pose tracking performance
in real-world scenarios.

\begin{table}[tbh]
\small
\centering
\resizebox{0.9\linewidth}{!}{
\begin{tabular}{c cc}
\toprule
\multirow{2}{*}{\textbf{Category}} & \multicolumn{2}{c}{\textbf{Per-part 6D Pose}} \\
\cline{2-3}
& Rotation Error ($^\circ$) & Translation Error ($m$) \\
\hline
Box & 4.8, 4.9 & 0.008, 0.008 \\
Stapler & 5.4, 5.8 & 0.010, 0.009 \\
Cutter  & 3.3, 3.3 & 0.008, 0.010 \\
Scissors  & 7.9, 8.7 & 0.007, 0.009 \\
Drawer  & 7.9, 7.9 & 0.025, 0.021 \\
\bottomrule
\end{tabular}}
\caption{Results on ReArt-Video Dataset.}
\label{tab:reart_exp}
\end{table}

\begin{table}[h!]
\scriptsize
\small
\centering
\resizebox{0.9\linewidth}{!}{
\begin{tabular}{ccccccc}
\toprule
\multicolumn{7}{c}{\textbf{Per-part Rotation Error ($^{\circ}$)}} \\
\hline
0.08 & 0.75 & 2.26 & 9.73 & 13.55 & 17.80 & 19.43 \\
\hline
\multicolumn{7}{c}{\textbf{Per-part Translation Error ($m$)}} \\
\hline
0.002 & 0.016 & 0.022 & 0.068 & 0.070 & 0.127 & 0.155 \\
\bottomrule
\end{tabular}}
\caption{Results on RobotArm-Videos Dataset.}
\label{tab:robotarm_exp}
\end{table}

\section{CONCLUSION}
This paper introduces PPF-Tracker, a framework for category-level articulated object pose tracking on the SE(3) manifold. Through a Quasi-Canonicalization strategy, the tracking task is reformulated as pose increment learning in Lie algebra space, ensuring stable and SE(3)-invariant motion modeling. To enhance accuracy and reduce drift, we adopt a dynamic keyframe selection mechanism based on geometric similarity and enforce structural consistency via kinematic constraints. Comprehensive evaluations on synthetic, semi-synthetic, and real-world datasets show that PPF-Tracker delivers state-of-the-art performance in accuracy, robustness, and generalization. We believe this work provides a principled and extensible foundation for articulated object tracking, with promising implications for robotics, embodied AI, and AR/VR applications.

\clearpage
\section*{Acknowledgements}
This work was supported by the National Natural Science Foundation of China under Contract 62471450.

\bibliography{aaai2026}

\end{document}